\pdfoutput=1

\documentclass[11pt]{article}

\usepackage{naacl2021}

\usepackage{times}
\usepackage{latexsym}

\usepackage[T1]{fontenc}

\usepackage[utf8]{inputenc}

\usepackage{microtype}

\usepackage{linguex}
\usepackage{booktabs}
\usepackage{graphicx}
\usepackage{colortbl}

\title{Is Incoherence Surprising? \\Targeted Evaluation of Coherence Prediction from Language Models}

\author{Anne Beyer 
		\And 
		Sharid Lo\'{a}iciga \\
Computational Linguistics, Department of Linguistics, University of Potsdam, Germany\\
\texttt{\{anne.beyer, loaicigasanchez, david.schlangen\}@uni-potsdam.de}
		\And 
		David Schlangen
}

\begin{document}

\maketitle
\begin{abstract}

Coherent discourse is distinguished from a mere collection of utterances by the satisfaction of a diverse set of constraints, for example choice of expression, logical relation between denoted events, and implicit compatibility with world-knowledge.
Do neural language models encode such constraints?
We design an extendable set of test suites addressing different aspects of discourse and dialogue coherence. Unlike most previous coherence evaluation studies, we address specific linguistic devices beyond sentence order perturbations, allowing for a more fine-grained analysis of what constitutes coherence and what neural models trained on a language modelling objective do encode. Extending the targeted evaluation paradigm for neural language models \citep{marvin-linzen-2018-targeted} to phenomena beyond syntax, we show that this paradigm is equally suited to evaluate linguistic qualities that contribute to the notion of coherence.
\end{abstract}

\section{Introduction}

Statistical models trained on large amounts of data using the language modelling objective (predicting words in context) have shown to pick up an intriguing amount of implicit knowledge about other tasks, for example syntactic knowledge \citep{warstadt-etal-2020-blimp, hu-etal-2020-systematic} or world knowledge \citep{trinh_simple_2019,tamborrino-etal-2020-pre}. They have also been shown to exhibit, within these tasks, interesting divergences from expectation and sensitivity to confounding factors (e.g. \citet{mccoy-etal-2019-right}). 

Inspired by the recently released SyntaxGym \citep{gauthier-etal-2020-SyntaxGym}, which enables specific and standardised evaluation of syntactic knowledge encoded in such models, we explore whether similar methods can be applied to the study of discourse knowledge or \textit{coherence}, i.e., constraints acting across sentence boundaries, as illustrated in \ref{ex:discourse} (where "\#" marks the less acceptable variant).

\ex. \label{ex:discourse}  
\a. \# The lone ranger rode off into the sunset. \textbf{Then} he \textit{jumped} on his horse.
\b. The lone ranger jumped on his horse. \textbf{Then} he \textit{rode} into the sunset. 

A common approach to coherence evaluation consists in shuffling the sentence order of a text, thereby creating incoherent text samples that need to be discriminated from the original \citep{Barzilay2008}. While this approach to creating incoherent test data is intuitive enough, recent studies suggest that it paints only a partial picture of what constitutes coherence \citep{lai-tetreault-2018-discourse,mohammadi-etal-2020-creation,pishdad-etal-2020-coherent}.
It does not pinpoint the qualities that make the shuffled text incoherent, it does not tell us which linguistic devices are at fault, emphasising the need to move beyond this technique. 
This paper aims to add to the growing body of research stressing the need for more qualitative evaluations of text coherence \citep{see-etal-2019-massively,mohammadi-etal-2020-creation,pishdad-etal-2020-coherent}. 

We design different test suites created semi-automatically from existing corpora. This eases the burden of creating them from scratch and ensures the inclusion of multiple genres, crucially including dialogue data. Each test suite addresses a hypothesis about an underlying linguistic device contributing to a text's coherence, i.e., choice of referring expressions, discourse connectives, and intention (speaker commitment).

Our contributions are the following:
We 
\begin{itemize}
    \item extend SyntaxGym to handle phenomena acting across sentence boundaries, but keep the general functionality to allow the use of both syntactic and coherence test suites,
    \item show that it is possible to evaluate dialogue models by extending \texttt{lm-zoo} (SyntaxGym's model repository), and
    \item present a first set of coherence test suites, each assessing a fine-grained and linguistically motivated element of coherence. 
\end{itemize}

Our work thus eliminates the need for adapting and gathering various benchmark datasets by providing an easily extensible coherence evaluation framework that allows the use of existing test suites and the design of new ones. At the moment, all of the test suites reported below are in English, but we come back to possible extensions in Section \ref{conclusion}.

Our results are mixed: To the extent that the test suites effectively capture coherence, the examined models are neither systematically incoherent nor coherent. We take this as support for our claim that more and better linguistically informed test suites are needed in oder to fully understand if neural models actually do capture genuine coherence. We expect to develop our work further, but at this point, our contribution is a systematic framework that will allow us to do just that.

The code to create our test suites can be found at \url{https://github.com/AnneBeyer/coherencegym}.

\section{Related Work}

\paragraph{SyntaxGym.} \citet{gauthier-etal-2020-SyntaxGym} develop a toolkit for targeted evaluation of language models on different syntactic phenomena. 
It is built on top of \texttt{lm-zoo},\footnote{\url{https://cpllab.github.io/lm-zoo/}} a repository of language models that each specify their corresponding function to extract token level surprisal values $s(t)$ from the language model's conditional token probabilities $p$.
\begin{equation}\label{eq:surp}
    s(t_i) = -log_2(p(t_i|t_0 \ldots t_{i-1}))
\end{equation}

Different syntactic phenomena can be evaluated by running models on different test suites. Each test suite contains items with minimally different conditions, focusing on the specific phenomenon. An example item for \textsc{number agreement} is given below.
\ex. \label{ex:sytaxgym}
\a. \textbf{condition name: match}\\
    \textit{region 1:} The woman \\
    \textit{region 2:} plays\\
    \textit{region 3:} the guitar\\
\b. \textbf{condition name: mismatch}\\
    \textit{region 1:} The woman \\
    \textit{region 2:} play\\
    \textit{region 3:} the guitar

Each test suite also contains a \textit{prediction} of the expected difference between conditions. Splitting the input into different regions makes it possible to measure the difference in model predictions at the token or phrase level. (e.g. \textit{region 2} in condition \textbf{mismatch} should be more surprising than \textit{region 2} in condition \textbf{match}).

\paragraph{Coherence.}
While the notion of syntactic acceptability is well studied from a linguistic point of view and in terms of neural language model representations \citep[\textit{inter alia}]{marvin-linzen-2018-targeted, warstadt-etal-2019-neural,warstadt-etal-2020-blimp,hu-etal-2020-systematic}, it remains less clear what neural models are capable of capturing when modelling language across sentence boundaries.

There exists a large body of work in linguistics regarding different notions of coherence, such as the influence of coreference \citep[\emph{inter alia}]{hobbs:1979, Barzilay2008}, Centering theory \citep{gjw:cent}, discourse structure \cite{manthom:rst, webber-etal-2003-anaphora}, and phenomena that connect utterances in dialogue, such as conversational maxims \citep{grice_logic_1975} or speaker interaction \citep{lascarides_agreement_2009}.

Many of these are also mentioned by coherence evaluation studies, nonetheless they mostly revert to the use of some form of sentence-order variations \citep{chen-etal-2019-evaluation, moon-etal-2019-unified, xu-etal-2019-cross,mesgar-etal-2020-dialogue}. 
While some progress has been made towards incorporating more linguistically motivated test sets \citep{chen-etal-2019-evaluation, mohammadi-etal-2020-creation, pishdad-etal-2020-coherent}, most evaluation studies focus on models trained specifically on coherence classification and prediction tasks.

\paragraph{Language models.} 
The recently proposed transformer language model \textsc{\textsc{GPT-2}} \citep{radford2019language} has been shown to perform very well on many downstream language tasks. \citet{see-etal-2019-massively} quantitatively evaluate  \textsc{\textsc{GPT-2}} as a language generator and find that it generally performs on par with a state-of-the-art neural story generation model. 
However, they also note that their automatic measures focus mostly on text diversity and stress the need for more qualitative evaluation methods for notions like text coherence.

\textsc{GPT-2} is also the basis of the recently proposed dialogue model \textsc{DialoGPT} \citep{zhang-etal-2020-dialogpt}, which is fine-tuned on conversational data from Reddit. 
\citet{mehri-eskenazi-2020-unsupervised} argue that \textsc{DialoGPT} encodes several notions of dialogue quality, including coherence. They manually create several positive and negative follow-up utterances for certain dialog qualities (e.g. ``Wow, that's interesting!" or ``I'm confused."). The likelihood of \textsc{DialoGPT} outputting either of them is then used to give an overall score per quality. The notion of dialogue coherence, although shown to be among the most important for predicting overall dialogue quality, is found to be one of the hardest to predict using this method. The authors attribute this to the fact that coherence (or the lack thereof) is seldom verbalised, so the model is not able to associate this notion with specific follow-up utterances. We take this a step back and evaluate the evaluator in order to get a better understanding of which notions of coherence are actually implicitly encoded in \textsc{DialoGPT}.

We test \textsc{\textsc{GPT-2}} and \textsc{DialoGPT} on different notions of discourse and dialogue coherence by evaluating them on specifically designed test suites building on the SyntaxGym methodology.

\section{From SyntaxGym to CoherenceGym: Querying Coherence Judgements and Creating Datasets}
We show that the methods implemented in SyntaxGym can also be applied to evaluate phenomena that go beyond a single sentence. 
SyntaxGym is based on the psycholinguistically motivated notion of surprisal, which they utilise to compare the scores assigned by a language model to specific regions in a minimal pair of sentences.
In our CoherenceGym setting, the regions of interest comprise larger chunks up to whole sentences. We calculate the models' token level surprisals and aggregate them over all tokens $t_1 \ldots t_n$ in the region $r$ of interest. As the continuations may differ in more than one token and can be of different lengths, we use the mean region surprisal.\footnote{This required a slight adaptation of \texttt{syntaxgym}, which is now part of the official implementation.}
\begin{equation}
    s_{mean}(r) = \frac{1}{n} \sum_{i=1}^{n} s(t_i)
\end{equation}

To create incoherent versions, we utilise several existing datasets and devise different modifications that target a concrete phenomenon. We also include some existing methods and resources in order to demonstrate that those can easily be integrated and to cover a wide range of phenomena, which are described in detail in Section \ref{sec:suites}.
We further add \textsc{DialoGPT} \citep{zhang-etal-2020-dialogpt} to the \texttt{lm-zoo} to show that the coherence test suites can also be used to evaluate dialogue models.\footnote{This implies some restrictions on compatibility though: All models should be able to predict discourse coherence phenomena, but only dialogue models need to additionally encode dialogue coherence.}

The Coherence Detection (CD) scores reported in Section \ref{sec:suites} measure the proportion of items for which each model met the prediction of each test suite, i.e., the prediction accuracy of whether the model found the incoherent version more surprising than the coherent counterpart.

\subsection{Models}
SyntaxGym is built as a wrapper on top of \texttt{lm-zoo}, a repository of language model Docker containers specifying the functions \texttt{tokenizer}, \texttt{unkify} and \texttt{get\_surprisals}. \textsc{GPT-2} (117M) \citep{radford2019language} is already included by the developers, based on the huggingface transformers library.\footnote{\url{https://huggingface.co/transformers/}} 
We use this version and add \textsc{DialoGPT} \citep{zhang-etal-2020-dialogpt}, which is built upon \textsc{GPT-2}, but further fine-tuned on Reddit data, in the same manner. As Reddit contains multi-person dialogues, the separator token is taken to denote speaker change.
Both models compute the next token probability based on the softmax output of the final linear layer of the decoder. Following the \texttt{get\_surprisals} function for \textsc{GPT-2}, we transform the token probabilities into surprisals as shown in Equation \ref{eq:surp}.

Each of the two models exist in different versions, depending on the number of parameters (embedding size, number of layers). 
For technical reasons, we used the small version of \textsc{GPT-2} (117M) and the medium version of \textsc{DialoGPT}(345M), so the two models are not directly comparable. 
As the aim of this study is to show that the surprisal based targeted evaluation paradigm is useful for coherence evaluation in general, we leave a detailed comparison of the impact of different model sizes to future work.

\section{Coherence Phenomena and Test Suites}
\label{sec:suites}

In this section, we describe the different coherence phenomena assessed by our test suites. For every test suite we first posit a hypothesis, which is coded into the suite's prediction section. Next, we describe the dataset and the manipulation applied to create incoherent samples that exhibit a violation of coherence regarding the specific phenomenon. %
Each subsection reports the results of the evaluated models on the respective test suite. As we evaluate models pre-trained on English data, our test suites are devised only in English as well.

The first three test suites are based on existing methods or test sets that we integrate into the framework. The following three test suites are newly created.

\subsection{Sentence Order Baseline Test Suite}\label{subsec:shuff}

\textit{Hypothesis: A coherent text is composed of an ordered set of sentences in a logical sequence; shuffling the sentences breaks the logical order and hence coherence.  Since sequentiality is central to the language modelling task, models successfully distinguish between both versions.} \\
This shuffling technique has been widely applied in the evaluation of coherence models \citep{Barzilay2008, chen-etal-2019-evaluation, moon-etal-2019-unified, xu-etal-2019-cross, mesgar-etal-2020-dialogue}.
We include it as baseline for our method, in order to contrast how more fine-grained notions of coherence compare to this broad approach. 

We use ROCStories \citep{mostafazadeh-etal-2016-corpus} and the \textsc{Persona-Chat} corpus \citep{zhang-etal-2018-personalizing} to evaluate sentence order for narration as well as dialogue data. The ROCStories corpus consists of coherent five-sentence stories which were gathered by employing crowdworkers and contain several temporal and causal relations between the sentences. To create the \textsc{Persona-Chat} corpus \citep{zhang-etal-2018-personalizing}, crowd sourced dialogue participants were assigned a persona in the form of descriptive natural language sentences and were asked to talk to each other impersonating their assigned persona. The dialogues contain at least 6 turns and we extract only the utterances and ignore the persona descriptions.

Two versions are created of both corpora:  \begin{enumerate}
    \item We shuffle all utterances and compare the aggregated overall surprisal for all tokens over all regions.
    \item We keep the last utterance fixed and shuffle only the context and compare the aggregated surprisal for the second region (cf. \ref{ex:shuff}).
\end{enumerate}

\ex. \label{ex:shuff}
\a. \textbf{condition name: original}\\
    \textit{region 1:} My friends all love to go to the club to dance. They think it's a lot of fun and always invite. I finally decided to tag along last Saturday. I danced terribly and broke a friend's toe.\\
    \textit{region 2:} The next weekend, I was asked to please stay home.
\b. \textbf{condition name: shuffled}\\
    \textit{region 1:} I finally decided to tag along last Saturday. I danced terribly and broke a friend's toe. My friends all love to go to the club to dance. They think it's a lot of fun and always invite. \\
    \textit{region 2:} The next weekend, I was asked to please stay home.

\paragraph{Results.}
\begin{table}[]
    \centering
    \tabcolsep=0.11cm
    \begin{tabular}{lcccc}
    \toprule
      & N\_all & N\_context& D\_all & D\_context\\
      \midrule
     \textsc{GPT-2} & $0.86$ &$0.50$& $0.96$ &$0.72$ \\
      \midrule
     \textsc{DialoGPT}& $0.78$ &$0.44$ & $0.86$ &$0.55$\\
      \midrule
     \textit{\#items} &$1871$& $1871$&$967$& $967$\\
    \bottomrule
    \end{tabular}
    \caption{CD scores on shuffling test suites. (N $=$ narration, D $=$ dialogue data, \textit{all} refers to the shuffling of all sentences, \textit{context} is based on comparing the surprisals of the last sentence with ordered or shuffled context.}
    \label{tab:shuff}
\end{table}

As Table \ref{tab:shuff} shows, shuffling is a good first indicator for detecting coherence on a global level, as the models perform quite well in the conditions where all sentences have been shuffled.\footnote{It is worth noting that by fine-tuning on user generated content, this ability decreases, which probably says more about Reddit than about\textsc{DialoGPT}, but as noted before, these results are not directly comparable as the models are of different sizes.}

On a local level (i.e., the influence that shuffling the context has on the following sentence), however, the ability to detect the manipulated sequence drops largely, even to or below chance.
A manual inspection of the data in the \textit{context} condition revealed that, in some cases, the final (non-moved) utterance (region 2) also can be judged as a coherent follow-up to the utterance shuffled into the final context position. 
This also reveals that shuffling does not always break coherence in the expected way due to the nature of natural language, thus highlighting the importance of a more thoughtful design of coherence test suites.

\subsection{Story Cloze Test Suite}
\textit{Hypothesis: Combining commonsense and discourse relations enables a model to detect a coherent from an incoherent ending of a given story.}\\
We use the same corpus as for the narration shuffling condition above, but keep the order intact. The Story Cloze test set \citep{mostafazadeh-etal-2016-corpus} contains an additional implausible ending to each story.
We use the annotated test set of the spring 2016 version and create items with different endings as exemplified in \ref{ex:roc}. 

\ex. \label{ex:roc}
\a. \textbf{condition name: original ending}\\
    \textit{region 1:} My friends all love to go to the club to dance. They think it's a lot of fun and always invite. I finally decided to tag along last Saturday. I danced terribly and broke a friend's toe.\\
    \textit{region 2:} The next weekend, I was asked to please stay home.
\b. \textbf{condition name: distractor ending}\\
    \textit{region 1:} My friends all love to go to the club to dance. They think it's a lot of fun and always invite. I finally decided to tag along last Saturday. I danced terribly and broke a friend's toe.\\
    \textit{region 2:} My friends decided to keep inviting me out as I am so much fun.
    
Calculating our CD score allows for a direct evaluation of language models without the need for training a classifier on top of the model representations.

\paragraph{Results.}
The first column in Table \ref{tab:roc_wino} displays the results on the Story Cloze test suite. 
While these results leave room for improvement, it is worth noting that they are on par or even outperform the models from the original paper, which mostly rely on semantic similarities between the context and the continuations.
However, we still do not learn which linguistic devices are responsible for the perception of coherence or incoherence of a given ending from this data. The following test suites are designed to investigate specific phenomena of coherence and models abilities to encode them in more detail.

\begin{table}[]
    \centering
    \begin{tabular}{lccc}
    \toprule
        &\multicolumn{1}{c}{Story Cloze}& \multicolumn{2}{c}{Winograd}\\
       &  & full & partial\\
      \midrule
     \textsc{GPT-2} & $0.61$& $0.53$& $0.59$\\
      \midrule
     \textsc{DialoGPT}& $0.57$ & $0.55$ & $0.57$\\
      \midrule
     \textit{\#items} &$1871$&$273$&$273$\\
     \bottomrule
    \end{tabular}
    \caption{CD scores on the Story Cloze and the Winograd test suites (\textit{full} is based on comparing the surprisals of the whole sequences, \textit{partial} only considers the regions following the inserted referent)}
    \label{tab:roc_wino}
\end{table}

\subsection{Winograd Schema Test Suite}
\textit{Hypothesis: Models are able to combine commonsense knowledge with pronoun resolution, thus they are able to distinguish the correct target from the distractor in Winograd Schema style sentences.}\\
This dataset was proposed by \citet{trinh_simple_2019} as has also been applied by \citet{radford2019language} for evaluating \textsc{GPT-2}'s commonsense knowledge. We reproduce the test suite in the following way:
\ex. \label{ex:winograd}
\a. \textbf{condition name: target}\\
    \textit{region 1:} The city councilmen refused the demonstrators a permit because \\
    \textit{region 2:} the city councilmen\\
    \textit{region 3:} feared violence.
\b. \textbf{condition name: distractor}\\
    \textit{region 1:} The city councilmen refused the demonstrators a permit because \\
    \textit{region 2:} the demonstrators\\
    \textit{region 3:} feared violence.

Following \citet{trinh_simple_2019} and \citet{radford2019language}, we compare the full version (comparing the mean surprisal over all tokens) and a partial version (comparing the surprisal for \textit{region 3}).

\paragraph{Results.}
The last two columns in Table \ref{tab:roc_wino} report the CD scores for the Winograd test suite.

As noted by \citet{trinh_simple_2019}, the difference in language model scores is more obvious in the region following the inserted correct or distracting entity. We are able to reproduce these results in our setting, which supports the applicability of the CoherenceGym approach. 
\citet{radford2019language} demonstrate that the performance on this task can be increased by adding more parameters to the model. We will inspect the impact of model sizes on the different test suites more closely in future work.

\subsection{Coreference Test Suite}
\textit{Hypothesis: Different referring expressions reflect both the accessibility and salience status of the entities being referred. For keeping in topic however, entities need only to be re-mentioned, regardless of their form. In this sense, language models are insensitive to the use of different referring expressions.}\\
In line with theories proposing an accessibility hierarchy that position pronouns requiring the highest level of accessibility and lexical noun phrases (indefinites and definites) the lowest level \citep[cf.]{Givon1983, Ariel2004}, we test whether language models capture a violation in the use of referring expressions according to their accessibility status.  

For this test suite, we work with the \textsc{arrau} corpus \citep{Uryupina-Et-al-2020-ARRAU}. In contrast to other coreference corpora, \textsc{arrau} is multi-genre --including news, dialogue and fiction texts-- and provides annotations for non-nominal anaphora such as discourse deixis. 

We extract coreferential chains whose mentions span consecutive sentences and with at least one pronominal mention. The test suites examples consist of minimal pairs \ref{ex:coreferenceex} where a same context sentence in \textit{region 1} containing the antecedent is followed by the sentence with the original pronoun re-mentioning the antecedent or by a manipulated sentence in which the pronoun is replaced by a repetition of the antecedent in \textit{region 2}. 

\ex. \label{ex:coreferenceex}
\a. \textbf{condition name: pronoun}\\
    \textit{region 1:} And there's a ladder coming out of the tree and there's a man at the top of the ladder \\
    \textit{region 2:} you can't see \textit{him} yet
\b. \textbf{condition name: repetition}\\
    \textit{region 1:} And there's a ladder coming out of the tree and there's a man at the top of the ladder \\
    \textit{region 2:} you can't see \textit{the man at the top of the ladder} yet

In keeping with the accessibility theory, we have replaced the indefinite marker \textit{a} with a definite \textit{the} in the \textbf{repetition} condition.

\paragraph{Results.} The results show that when presented with a new lexical entity, neither model has a clear preference for a pronominal re-mention of the entity (Table \ref{tab:entity}). The very nature of the language model will drive it to topic continuity, as it is designed to generate tokens based on a previous history. However, this does not automatically ensures cohesion. Both pronominalisation and repetition represent cohesive ties to the previous context recoverable from surface cues. The difference is that the first involves a stronger link with the context, licensing the use of the pronoun, which the models evaluated here fail to pick up.

\begin{table}[]
    \centering
    \begin{tabular}{lcccc}
    \toprule
       & \textsc{wsj} & \textsc{vpc} & Dialogue & Fiction \\
      \midrule
     \textsc{GPT-2} & $0.53$& $0.56$& $0.47$& $0.42$ \\
      \midrule
     \textsc{DialoGPT}& $0.44$ & $0.51$&$0.47$&$0.36$ \\
      \midrule
     \textit{\#items} &$512$&$75$&$68$&$98$\\
     \bottomrule
    \end{tabular}
    \caption{CD score results on entity re-mention test suite. \textsc{wsj} and \textsc{vpc} refer to the News portion of the \textsc{arrau} corpus.}
    \label{tab:entity}
\end{table}

\subsection{Explicit Connectives Test Suite}
\textit{Hypothesis: Meaning is constructed by building a representation for each new sentence based on the content of the previous sentences, and a first level of the coherence between two segments is embodied by explicit connectives. Hence, an inappropriate connective between two segments will yield a content gap. Sensitivity to content-meaning implies then sensitivity to a change in explicit connectives.}\\
For this exercise, we work with Disco-Annotation \citep{Popescu-Belis-LREC-2012}, a corpus of segments from the Europarl corpus \citep{Koehn2005} annotated with discourse connective senses.\footnote{Europarl segments are either very long sentences formed by several clauses or by 2-3 sentences clustered together, as a product of the sentence alignment process.} Eight discourse connectives are annotated in the corpus (\textit{as, although, though, while, since, yet, however, meanwhile}), with one of five possible senses (\textit{contrast, concession, causal, temporal, comparison}). We excluded all examples where the connective is in a segment initial position, since the previous segment is not provided, a setting incompatible with our constraints. This removed all examples of \textit{meanwhile}. A minimal pair is created from each segment \ref{ex:connectives}, where all the tokens up to the connective are used as context, followed by the original connective or another connective from the set, and the continuation of the segment.

\ex. \label{ex:connectives}
\a. \textbf{condition name: original}\\
    \textit{region 1:} We share the widespread outrage at its attitude to history, in particular World War II, but also its policies on enlargement, on immigration, on race and its attitude to the European Union itself. We were also outraged,\\
    \textit{region 2:} \textit{however}\\
    \textit{region 3:} , at the tolerance of the left for the tyranny, the terror and the excesses of the former USSR.
\b. \textbf{condition name: manipulated}\\
    \textit{region 1:} We share the widespread outrage at its attitude to history, in particular World War II, but also its policies on enlargement, on immigration, on race and its attitude to the European Union itself. We were also outraged,\\
    \textit{region 2:} \textit{since}\\
    \textit{region 3:} , at the tolerance of the left for the tyranny, the terror and the excesses of the former USSR.
 
Some connectives may have the same sense depending on the specific context in which they appear \citep{Stede2012, webber-etal-2003-anaphora}, for instance both \textit{since} and \textit{while} may bear a \textit{temporal} interpretation. On that account, we expect that a replacement with a different connective bearing a different sense leads to \textbf{region 3} being more surprising than a different connective able to have the same sense.  

\paragraph{Results.} Not all relations captured by the connectives are equally difficult, producing high variability in the scores, as shown in Table \ref{tab:connectivesresults}. While temporal senses seem to be relatively unproblematic (scores about 0.85 on average, \textsc{\textsc{GPT-2}}), `contrast', `concession' and in particular `causal' senses are more difficult to distinguish (\textit{since\_causal} and \textit{as\_causal} have averages of 0.66 and 0.52 respectively). 

The results for \textit{as} present an interesting contrast. This connective can also be used as a preposition. When the connectives with this particular sense are replaced, the models do not have any trouble recognising the original from the manipulated sentence, as suggested by the systematic high scores obtained, between 0.96 and 0.99. In most other senses, however, scores plummet as low as 0.28. We observe a similar pattern for \textit{yet} when used as an adverb in the \textsc{DialogPT} model.

\begin{table*}
\resizebox{\linewidth}{!}{%
\begin{tabular}{l*{7}{r}|*{7}{r}}
\toprule
& \multicolumn{7}{c}{\textsc{\textsc{GPT-2}}} & \multicolumn{7}{c}{\textsc{dialogpt}} \\
& \multicolumn{7}{c}{Connective used in manipulation} & \multicolumn{7}{c}{Connective used in manipulation} \\
\midrule
\textsc{connective sense} &	although &	as&	however&	since	&though&	while&	yet		&although	&as&	however	&since	&though&	while&	yet \\
although\_concession &	--&	$0.92$ &	$0.92$ &	$0.857$ &	$0.84$ &	$0.86$ &	$0.90$ &		-- &	$0.88$ &	$0.83$ &	$0.82$ &	$0.76$ &	$0.76$ &	$0.89$ \\
although\_contrast &	-- &	$1.00$ &	$0.86$ &	$0.86$ &	$\textbf{0.43}$ &	$1.00$ &	$0.86$ &		-- &	$1.00$ &	$0.86$ &	$1.00$ &	$0.93$ &	$0.79$ &	$1.00$ \\
\midrule															
as\_causal &	$\textbf{0.44}$ &	-- &	$0.80$ &	$\textbf{0.28}$ &	$0.64$ &	$0.72$ &	$0.80$ &		$0.64$ &	--	& $0.68$ &	$\textbf{0.40}$ &	$0.84$ &	$0.80$ &	$0.88$ \\
as\_comparison &	$0.96$ &	-- &	$0.95$ &	$0.96$ &	$0.94$ &	$0.97$ &	$0.97$ &		$0.86$ &	-- &	$0.89$ &	$0.86$ &	$0.93$ &	$0.87$ &	$0.92$ \\
as\_concession &	$\textbf{0.33}$ &	-- &	$0.67$ &	$0.67$ &	$\textbf{0.33}$ &	$1.00$ &	$1.00$ &		$0.67$ &	-- &	$0.67$ &	$0.67$ &	$1.00$ &	$1.00$ &	$0.67$ \\
as\_\textsc{preposition}&\cellcolor{yellow}	$0.99$ &\cellcolor{yellow}	-- &\cellcolor{yellow}	$0.99$ &\cellcolor{yellow}	$0.98$ &\cellcolor{yellow}	$0.99$ &\cellcolor{yellow}	$0.99$ &\cellcolor{yellow}	$0.99$ &\cellcolor{yellow}		$0.96$ &\cellcolor{yellow}	-- &\cellcolor{yellow}	$0.97$ &\cellcolor{yellow}	$0.97$ &\cellcolor{yellow}	$0.97$ &\cellcolor{yellow}	$0.97$ &\cellcolor{yellow}	$0.98$ \\
as\_temporal&	$0.95$ &	-- &	$0.95$ &	$0.86$ &	$1.00$ &	$0.81$ &	$0.95$ &		$0.86$ &	-- &	$1.00$ &	$0.86$ &	$1.00$ &	$0.76$ &	$1.00$ \\
\midrule												
however\_concession	& $0.70$ &	$0.90$ &	-- &	$0.86$ &	$0.63$ &	$0.80$ &	$0.64$ &		$0.58$ &	$0.79$ &	--	 & $0.79$ &	$0.71$ &	$0.71$ &	$0.53$ \\
however\_contrast &	$0.67$ &	$0.89$ &	-- &	$0.89$ &	$\textbf{0.33}$ &	$0.89$ &	$0.67$ &		$0.67$ &	$0.89$ &	-- &	$0.78$ &	$0.56$ &	$0.67$ &	$0.56$ \\
\midrule															
since\_causal &	$0.61$ &	$0.78$ &	$0.83$ &	-- &	$0.72$ &	$0.79$ &	$0.93$ &		$0.66$ &	$0.82$ &	$0.74$ &	--&	$0.83$ &	$0.79$ &	$0.89$ \\
since\_temporal-causal &	$1.00$ &	$0.83$ &	$1.00$ &	-- &	$1.00$ &	$1.00$ &	$1.00$ &		$0.67$ &	$1.00$ &	$1.00$ &	-- &	$0.83$ &	$0.83$ &	$1.00$ \\
since\_temporal &	$0.96$ &	$0.97$ &	$0.96$ &	-- &	$0.94$ &	$0.97$ &	$0.98$ &		$0.95$ &	$0.97$ &	$0.95$ &	-- &	$0.95$ &	$0.92$ &	$0.95$ \\
\midrule															
though\_concession &	$\textbf{0.43}$ &	$0.82$ &	$0.79$ &	$0.87$ &	--&	$0.78$ &	$0.90$ &		$\textbf{0.37}$ &	$0.87$ &	$0.76$ &	$0.82$ &	-- &	$0.78$ &	$0.79$ \\
though\_contrast &	$0.56$ &	$0.84$ &	$0.84$ &	$0.88$ &	--&	$0.88$ &	$0.80$ &		$\textbf{0.41}$ &	$0.75$ &	$0.59$ &	$0.77$ &	-- &	$0.72$ &	$0.83$ \\
\midrule															
		
while\_concession &	$\textbf{0.46}$ &	$1.00$ &	$1.00$ &	$0.96$ &	$0.78$ &	--&	$0.98$ &		$0.57$ &	$0.96$ &	$0.87$ &	$0.89$ &	$0.76$ &	-- &	$0.93$ \\
while\_contrast &	$0.78$ &	$0.93$ &	$0.93$ &	$0.85$ &	$0.81$ &	--&	$0.81$ &		$0.81$ &	$0.93$ &	$0.81$ &	$0.85$ &	$0.96$ &	--&	$0.81$ \\
while\_temporal-causal &	$0.90$ &	$0.80$ &	$0.90$ &	$0.70$ &	$0.80$ &	--&	$0.80$ &		$0.80$ &	$0.80$ &	$1.00$ &	$0.70$ &	$0.90$ &	--&	$1.00$ \\
while\_temporal-contrast &	$0.73$ &	$0.90$ &	$0.90$ &	$0.73$ &	$0.77$ &	--&	$0.81$ &		$0.67$ &	$0.81$ &	$0.81$ &	$0.85$ &	$0.88$ &	--&	$0.81$ \\
while\_temporal &	$0.57$ &	$0.57$ &	$0.71$ &	$0.86$ &	$0.71$ &	--&	$0.86$ &		$0.86$ &	$0.71$ &	$0.86$ &	$0.86$ &	$1.00$ &	--&	$1.00$ \\
\midrule															
yet\_ADV &\cellcolor{yellow}	$0.95$ &\cellcolor{yellow}	$0.98$ &\cellcolor{yellow}	$0.82$ &\cellcolor{yellow}	$0.93$ &\cellcolor{yellow}	$0.93$ &\cellcolor{yellow}	$0.98$ &\cellcolor{yellow}	--	&\cellcolor{yellow}	$0.92$ &\cellcolor{yellow}	$0.96$ &\cellcolor{yellow}	$0.85$ &\cellcolor{yellow}	$0.92$ &\cellcolor{yellow}	$0.94$ &\cellcolor{yellow}	$0.97$ &\cellcolor{yellow}	-- \\
yet\_concession	 & $0.92$ &	$0.95$ &	$0.95$ &	$0.92$ &	$0.97$ &	$0.95$ &	--	&	$0.59$ &	$0.90$ &	$0.72$ &	$0.79$ &	$0.74$ &	$0.85$ &	-- \\
yet\_contrast & 	$0.92$ &	$0.88$ &	$0.88$ &	$0.92$ &	$0.92$ &	$0.79$ &	--	&	$0.67$ &	$0.96$ &	$0.75$ &	$0.88$ &	$0.79$ &	$0.88$ &	-- \\
\bottomrule
\end{tabular}%
}
\caption{CD scores on explicit connectives test suite. The first column list all the connective senses from Disco-Annotation. Scores below 0.50 are boldfaced, while the \textsc{preposition} and \textsc{adverb} senses are highlighted in yellow.}\label{tab:connectivesresults}
\end{table*}

\subsection{Speaker Commitment Test Suite}
\textit{Hypothesis: While it is possible for different speakers to have different opinions, speakers should not contradict themselves.}
This test suite targets the notion of speaker commitment in dialogue models. The test suite is created automatically based on the DialogueNLI corpus \citep{welleck-etal-2019-dialogue}, which contains pairs of utterances annotated as contradiction, entailment or neutral. 
The sentence pairs are extracted from the \textsc{persona-chat} corpus introduced in Section \ref{subsec:shuff}. The sentences can either be part of the conversation or the persona descriptions. We extract the contradicting sentence pairs from the human verified test set, and create two conditions for each utterance pair, as illustrated below:

\ex. \label{ex:dialog}
\a. \textbf{condition name: speaker change}\\
    \textit{region 1:} since the beginning of the year, i am a nurse. [SEP] \\
    \textit{region 2:} i am a kindergarten teacher.
\b. \textbf{condition name: same speaker}\\
    \textit{region 1:} since the beginning of the year, i am a nurse. \\
    \textit{region 2:} i am a kindergarten teacher.

In the first condition, we simulate a speaker change by introducing a \textit{[SEP]} token (which is converted to the tokenizer's separator token internally) in the dialogue history, whereas in the second condition the continuation is uttered by the same speaker as the context.

A model that is encoding some notion of speaker commitment should find the second utterance more surprising if no speaker change occurred.

As non-dialogue language models do not encode the notion of speaker change, this test suite only yields relevant results for dialogue models.

\paragraph{Results.}
\begin{table}[]
    \centering
    \tabcolsep=0.11cm
    \begin{tabular}{lc}
    \toprule
      & contradiction\\
      \midrule
     \textsc{DialoGPT}& 0.59\\
      \midrule
     \textit{\#items} &$4041$\\
    \bottomrule
    \end{tabular}
    \caption{CD score for speaker commitment test suite.}
    \label{tab:commit}
\end{table}

\textsc{DialoGPT} shows a tendency towards finding contradictions within the same speaker more surprising. A manual inspection of the data revealed that even though we use the human verified test set, there are quite some instances where the implications are not as clear, for example in the following two sentence pairs:
\ex.
\a. "my nurse skills come in handy when i volunteer."\\
"i am a kindergarten teacher."
\b. "i love art and want to be a famous artist."\\
"i am a kindergarten teacher."

This highlights the importance of quality over quantity. In future work, we will inspect this phenomenon more closely and combine the selection of items with human evaluation, to gain a better understanding of how the notion of speaker commitment is and can be encoded in neural dialogue models.

\section{Conclusions}
\label{conclusion}
We revisit the targeted evaluation paradigm and create test suites focusing on specific coherence phenomena. Each test suite contains minimal pairs of sequences that illustrate a specific component of coherence.

We evaluate two transformer models for language and dialogue modelling based on the token level surprisal scores they assign to the coherent and incoherent versions. Extending the existing SyntaxGym toolkit, we evaluate \textsc{GPT-2} and \textsc{DialoGPT} on our newly designed test suites on entity re-mention, explicit discourse connectives and speaker commitment in dialogue. Existing test sets are also integrated easily, which we demonstrate for sentence order detection, Story Cloze and Winograd Schema resolution tasks. 
Our results support previous work suggesting that the notion of coherence encoded in neural language models is more nuanced than the sentence order discrimination task can reflect. 

The mixed results we get, with some manipulations (e.g. the different sense connective substitutions) easily being spotted by the tested models and others (e.g. how to re-mention entities, or speaker contradictions) posing to be more difficult, point to the value of such targeted evaluation, which eventually might help in pointing towards where the introduction of different inductive biases could increase a model's performance.

In this study, we focus on the English language. However, our approach is not inherently designed for English alone. While \texttt{lm-zoo} only contains English language models at the moment, other language models can be added easily. The shuffling perturbations can be applied to any corpus. Our other test suites are based on available annotated corpora, which require some familiarity with the language, but can in principle be applied in a similar fashion to resources in other languages, such as the Potsdam Commentary Corpus \citep{bourgonje-stede-2020-potsdam} for German connectives, for example. We leave a multilingual extension of our framework for future work.

Our next efforts will focus on adding more language and dialogue models to determine the impact of different model architectures and sizes. Building additional test suites in order to capture a more thorough notion of coherence is also among our priorities. Last, we plan to collect human judgements to evaluate our coherence manipulations more closely and to create an upper bound for what we can expect from neural models.

\section*{Acknowledgements}
We thank Johann Seltmann and Jon Gauthier for their help with augmenting \texttt{lm-zoo} and \texttt{syntaxgym}.
We also thank the anonymous reviewers for their valuable feedback. 
This work was funded by the German Research Foundation (Deutsche Forschungsgemeinschaft, DFG) – Project ID 317633480 – SFB 1287.
\bibliography{anthology,custom}

\begin{thebibliography}{36}
\expandafter\ifx\csname natexlab\endcsname\relax\def\natexlab#1{#1}\fi

\bibitem[{Ariel(2004)}]{Ariel2004}
Mira Ariel. 2004.
\newblock Accessibility marking: Discourse functions, discourse profiles, and
  processing cues.
\newblock \emph{Discourse Processes}, 37(2):91–116.

\bibitem[{Barzilay and Lapata(2008)}]{Barzilay2008}
Regina Barzilay and Mirella Lapata. 2008.
\newblock {Modeling Local Coherence: An Entity-Based Approach}.
\newblock \emph{Computational Linguistics}, 34(1):1--34.

\bibitem[{Bourgonje and Stede(2020)}]{bourgonje-stede-2020-potsdam}
Peter Bourgonje and Manfred Stede. 2020.
\newblock \href {https://www.aclweb.org/anthology/2020.lrec-1.133} {The
  {P}otsdam commentary corpus 2.2: Extending annotations for shallow discourse
  parsing}.
\newblock In \emph{Proceedings of the 12th Language Resources and Evaluation
  Conference}, pages 1061--1066, Marseille, France. European Language Resources
  Association.

\bibitem[{Chen et~al.(2019)Chen, Chu, and Gimpel}]{chen-etal-2019-evaluation}
Mingda Chen, Zewei Chu, and Kevin Gimpel. 2019.
\newblock \href {https://doi.org/10.18653/v1/D19-1060} {Evaluation benchmarks
  and learning criteria for discourse-aware sentence representations}.
\newblock In \emph{Proceedings of the 2019 Conference on Empirical Methods in
  Natural Language Processing and the 9th International Joint Conference on
  Natural Language Processing (EMNLP-IJCNLP)}, pages 649--662, Hong Kong,
  China. Association for Computational Linguistics.

\bibitem[{Gauthier et~al.(2020)Gauthier, Hu, Wilcox, Qian, and
  Levy}]{gauthier-etal-2020-SyntaxGym}
Jon Gauthier, Jennifer Hu, Ethan Wilcox, Peng Qian, and Roger Levy. 2020.
\newblock \href {https://doi.org/10.18653/v1/2020.acl-demos.10} {{S}yntax{G}ym:
  An online platform for targeted evaluation of language models}.
\newblock In \emph{Proceedings of the 58th Annual Meeting of the Association
  for Computational Linguistics: System Demonstrations}, pages 70--76, Online.
  Association for Computational Linguistics.

\bibitem[{Givón(1983)}]{Givon1983}
Thomas Givón. 1983.
\newblock \emph{Topic Continuity in Discourse: A Quantitative Cross-Language
  Study}.
\newblock John Benjamin, Amsterdam.

\bibitem[{Grice(1975)}]{grice_logic_1975}
Herbert~P Grice. 1975.
\newblock Logic and conversation.
\newblock In \emph{Speech acts}, pages 41--58. Brill.

\bibitem[{Grosz et~al.(1995)Grosz, Joshi, and Weinstein}]{gjw:cent}
Barbara Grosz, Aravind~K. Joshi, and Scott Weinstein. 1995.
\newblock Centering: A framework for modelling the local coherence of
  discourse.
\newblock \emph{Computational Linguistics}, 21(2):203--225.

\bibitem[{Hobbs(1979)}]{hobbs:1979}
Jerry~R Hobbs. 1979.
\newblock {Coherence and Coreference}.
\newblock \emph{Cognitive Science}, 3(1):67--90.

\bibitem[{Hu et~al.(2020)Hu, Gauthier, Qian, Wilcox, and
  Levy}]{hu-etal-2020-systematic}
Jennifer Hu, Jon Gauthier, Peng Qian, Ethan Wilcox, and Roger Levy. 2020.
\newblock \href {https://doi.org/10.18653/v1/2020.acl-main.158} {A systematic
  assessment of syntactic generalization in neural language models}.
\newblock In \emph{Proceedings of the 58th Annual Meeting of the Association
  for Computational Linguistics}, pages 1725--1744, Online. Association for
  Computational Linguistics.

\bibitem[{Koehn(2005)}]{Koehn2005}
Philipp Koehn. 2005.
\newblock {Europarl}: A parallel corpus for statistical machine translation.
\newblock In \emph{Proceedings of the 10th Machine Translation Summit}, MT
  Summit X, pages 79--86, Phuket, Thailand.

\bibitem[{Lai and Tetreault(2018)}]{lai-tetreault-2018-discourse}
Alice Lai and Joel Tetreault. 2018.
\newblock \href {https://doi.org/10.18653/v1/W18-5023} {Discourse coherence in
  the wild: A dataset, evaluation and methods}.
\newblock In \emph{Proceedings of the 19th Annual {SIG}dial Meeting on
  Discourse and Dialogue}, pages 214--223, Melbourne, Australia. Association
  for Computational Linguistics.

\bibitem[{Lascarides and Asher(2009)}]{lascarides_agreement_2009}
A.~Lascarides and N.~Asher. 2009.
\newblock \href {https://doi.org/10.1093/jos/ffn013} {Agreement, {Disputes} and
  {Commitments} in {Dialogue}}.
\newblock \emph{Journal of Semantics}, 26(2):109--158.

\bibitem[{Mann and Thompson(1987)}]{manthom:rst}
William~C. Mann and Sandra~A. Thompson. 1987.
\newblock Rhetorical structure theory: A theory of text organization.
\newblock In Livia Polanyi, editor, \emph{The Structure of Discourse}. Ablex
  Publishing Corporation, Norwood, N.J.

\bibitem[{Marvin and Linzen(2018)}]{marvin-linzen-2018-targeted}
Rebecca Marvin and Tal Linzen. 2018.
\newblock \href {https://doi.org/10.18653/v1/D18-1151} {Targeted syntactic
  evaluation of language models}.
\newblock In \emph{Proceedings of the 2018 Conference on Empirical Methods in
  Natural Language Processing}, pages 1192--1202, Brussels, Belgium.
  Association for Computational Linguistics.

\bibitem[{McCoy et~al.(2019)McCoy, Pavlick, and Linzen}]{mccoy-etal-2019-right}
Tom McCoy, Ellie Pavlick, and Tal Linzen. 2019.
\newblock \href {https://doi.org/10.18653/v1/P19-1334} {Right for the wrong
  reasons: Diagnosing syntactic heuristics in natural language inference}.
\newblock In \emph{Proceedings of the 57th Annual Meeting of the Association
  for Computational Linguistics}, pages 3428--3448, Florence, Italy.
  Association for Computational Linguistics.

\bibitem[{Mehri and Eskenazi(2020)}]{mehri-eskenazi-2020-unsupervised}
Shikib Mehri and Maxine Eskenazi. 2020.
\newblock \href {https://www.aclweb.org/anthology/2020.sigdial-1.28}
  {Unsupervised evaluation of interactive dialog with {D}ialo{GPT}}.
\newblock In \emph{Proceedings of the 21th Annual Meeting of the Special
  Interest Group on Discourse and Dialogue}, pages 225--235, 1st virtual
  meeting. Association for Computational Linguistics.

\bibitem[{Mesgar et~al.(2020)Mesgar, B{\"u}cker, and
  Gurevych}]{mesgar-etal-2020-dialogue}
Mohsen Mesgar, Sebastian B{\"u}cker, and Iryna Gurevych. 2020.
\newblock \href {https://doi.org/10.18653/v1/2020.acl-main.133} {Dialogue
  coherence assessment without explicit dialogue act labels}.
\newblock In \emph{Proceedings of the 58th Annual Meeting of the Association
  for Computational Linguistics}, pages 1439--1450, Online. Association for
  Computational Linguistics.

\bibitem[{Mohammadi et~al.(2020)Mohammadi, Beiko, and
  Kosseim}]{mohammadi-etal-2020-creation}
Elham Mohammadi, Timothe Beiko, and Leila Kosseim. 2020.
\newblock \href {https://www.aclweb.org/anthology/2020.lrec-1.134} {On the
  creation of a corpus for coherence evaluation of discursive units}.
\newblock In \emph{Proceedings of the 12th Language Resources and Evaluation
  Conference}, pages 1067--1072, Marseille, France. European Language Resources
  Association.

\bibitem[{Moon et~al.(2019)Moon, Mohiuddin, Joty, and
  Xu}]{moon-etal-2019-unified}
Han~Cheol Moon, Tasnim Mohiuddin, Shafiq Joty, and Chi Xu. 2019.
\newblock \href {https://doi.org/10.18653/v1/D19-1231} {A unified neural
  coherence model}.
\newblock In \emph{Proceedings of the 2019 Conference on Empirical Methods in
  Natural Language Processing and the 9th International Joint Conference on
  Natural Language Processing (EMNLP-IJCNLP)}, pages 2262--2272, Hong Kong,
  China. Association for Computational Linguistics.

\bibitem[{Mostafazadeh et~al.(2016)Mostafazadeh, Chambers, He, Parikh, Batra,
  Vanderwende, Kohli, and Allen}]{mostafazadeh-etal-2016-corpus}
Nasrin Mostafazadeh, Nathanael Chambers, Xiaodong He, Devi Parikh, Dhruv Batra,
  Lucy Vanderwende, Pushmeet Kohli, and James Allen. 2016.
\newblock \href {https://doi.org/10.18653/v1/N16-1098} {A corpus and cloze
  evaluation for deeper understanding of commonsense stories}.
\newblock In \emph{Proceedings of the 2016 Conference of the North {A}merican
  Chapter of the Association for Computational Linguistics: Human Language
  Technologies}, pages 839--849, San Diego, California. Association for
  Computational Linguistics.

\bibitem[{Pishdad et~al.(2020)Pishdad, Fancellu, Zhang, and
  Fazly}]{pishdad-etal-2020-coherent}
Leila Pishdad, Federico Fancellu, Ran Zhang, and Afsaneh Fazly. 2020.
\newblock \href {https://doi.org/10.18653/v1/2020.coling-main.539} {How
  coherent are neural models of coherence?}
\newblock In \emph{Proceedings of the 28th International Conference on
  Computational Linguistics}, pages 6126--6138, Barcelona, Spain (Online).
  International Committee on Computational Linguistics.

\bibitem[{Popescu-Belis et~al.(2012)Popescu-Belis, Meyer, Liyanapathirana,
  Cartoni, and Zufferey}]{Popescu-Belis-LREC-2012}
Andrei Popescu-Belis, Thomas Meyer, Jeevanthi Liyanapathirana, Bruno Cartoni,
  and Sandrine Zufferey. 2012.
\newblock {D}iscourse-level {A}nnotation over {E}uroparl for {M}achine
  {T}ranslation: {C}onnectives and {P}ronouns.
\newblock In \emph{Proceedings of the eighth international conference on
  Language Resources and Evaluation ({LREC})}, Istanbul, Turkey.

\bibitem[{Radford et~al.(2019)Radford, Wu, Child, Luan, Amodei, and
  Sutskever}]{radford2019language}
Alec Radford, Jeff Wu, Rewon Child, David Luan, Dario Amodei, and Ilya
  Sutskever. 2019.
\newblock Language models are unsupervised multitask learners.
\newblock \emph{Technical report, {OpenAI}}.

\bibitem[{See et~al.(2019)See, Pappu, Saxena, Yerukola, and
  Manning}]{see-etal-2019-massively}
Abigail See, Aneesh Pappu, Rohun Saxena, Akhila Yerukola, and Christopher~D.
  Manning. 2019.
\newblock \href {https://doi.org/10.18653/v1/K19-1079} {Do massively pretrained
  language models make better storytellers?}
\newblock In \emph{Proceedings of the 23rd Conference on Computational Natural
  Language Learning (CoNLL)}, pages 843--861, Hong Kong, China. Association for
  Computational Linguistics.

\bibitem[{Stede(2012)}]{Stede2012}
Manfred Stede. 2012.
\newblock \emph{Disourse Processing}.
\newblock Morgan and Claypool Publishers, Toronto.

\bibitem[{Tamborrino et~al.(2020)Tamborrino, Pellican{\`o}, Pannier, Voitot,
  and Naudin}]{tamborrino-etal-2020-pre}
Alexandre Tamborrino, Nicola Pellican{\`o}, Baptiste Pannier, Pascal Voitot,
  and Louise Naudin. 2020.
\newblock \href {https://doi.org/10.18653/v1/2020.acl-main.357} {Pre-training
  is (almost) all you need: An application to commonsense reasoning}.
\newblock In \emph{Proceedings of the 58th Annual Meeting of the Association
  for Computational Linguistics}, pages 3878--3887, Online. Association for
  Computational Linguistics.

\bibitem[{Trinh and Le(2019)}]{trinh_simple_2019}
Trieu~H. Trinh and Quoc~V. Le. 2019.
\newblock \href {http://arxiv.org/abs/1806.02847} {A {Simple} {Method} for
  {Commonsense} {Reasoning}}.
\newblock \emph{arXiv:1806.02847 [cs]}.

\bibitem[{Uryupina et~al.(2020)Uryupina, Artstein, Bristot, Cavicchio, Delogu,
  Rodriguez, and Poesio}]{Uryupina-Et-al-2020-ARRAU}
Olga Uryupina, Ron Artstein, Antonella Bristot, Federica Cavicchio, Francesca
  Delogu, Kepa Rodriguez, and Massimo Poesio. 2020.
\newblock Annotating a broad range of anaphoric phenomena, in multiple genres:
  the {ARRAU} corpus.
\newblock \emph{Natural Language Engineering}, 26:95--128.

\bibitem[{Warstadt et~al.(2020)Warstadt, Parrish, Liu, Mohananey, Peng, Wang,
  and Bowman}]{warstadt-etal-2020-blimp}
Alex Warstadt, Alicia Parrish, Haokun Liu, Anhad Mohananey, Wei Peng, Sheng-Fu
  Wang, and Samuel~R. Bowman. 2020.
\newblock \href {https://doi.org/10.1162/tacl_a_00321} {{BL}i{MP}: The
  benchmark of linguistic minimal pairs for {E}nglish}.
\newblock \emph{Transactions of the Association for Computational Linguistics},
  8:377--392.

\bibitem[{Warstadt et~al.(2019)Warstadt, Singh, and
  Bowman}]{warstadt-etal-2019-neural}
Alex Warstadt, Amanpreet Singh, and Samuel~R. Bowman. 2019.
\newblock \href {https://doi.org/10.1162/tacl_a_00290} {Neural network
  acceptability judgments}.
\newblock \emph{Transactions of the Association for Computational Linguistics},
  7:625--641.

\bibitem[{Webber et~al.(2003)Webber, Stone, Joshi, and
  Knott}]{webber-etal-2003-anaphora}
Bonnie Webber, Matthew Stone, Aravind Joshi, and Alistair Knott. 2003.
\newblock \href {https://doi.org/10.1162/089120103322753347} {Anaphora and
  discourse structure}.
\newblock \emph{Computational Linguistics}, 29(4):545--587.

\bibitem[{Welleck et~al.(2019)Welleck, Weston, Szlam, and
  Cho}]{welleck-etal-2019-dialogue}
Sean Welleck, Jason Weston, Arthur Szlam, and Kyunghyun Cho. 2019.
\newblock \href {https://doi.org/10.18653/v1/P19-1363} {Dialogue natural
  language inference}.
\newblock In \emph{Proceedings of the 57th Annual Meeting of the Association
  for Computational Linguistics}, pages 3731--3741, Florence, Italy.
  Association for Computational Linguistics.

\bibitem[{Xu et~al.(2019)Xu, Saghir, Kang, Long, Bose, Cao, and
  Cheung}]{xu-etal-2019-cross}
Peng Xu, Hamidreza Saghir, Jin~Sung Kang, Teng Long, Avishek~Joey Bose,
  Yanshuai Cao, and Jackie Chi~Kit Cheung. 2019.
\newblock \href {https://doi.org/10.18653/v1/P19-1067} {A cross-domain
  transferable neural coherence model}.
\newblock In \emph{Proceedings of the 57th Annual Meeting of the Association
  for Computational Linguistics}, pages 678--687, Florence, Italy. Association
  for Computational Linguistics.

\bibitem[{Zhang et~al.(2018)Zhang, Dinan, Urbanek, Szlam, Kiela, and
  Weston}]{zhang-etal-2018-personalizing}
Saizheng Zhang, Emily Dinan, Jack Urbanek, Arthur Szlam, Douwe Kiela, and Jason
  Weston. 2018.
\newblock \href {https://doi.org/10.18653/v1/P18-1205} {Personalizing dialogue
  agents: {I} have a dog, do you have pets too?}
\newblock In \emph{Proceedings of the 56th Annual Meeting of the Association
  for Computational Linguistics (Volume 1: Long Papers)}, pages 2204--2213,
  Melbourne, Australia. Association for Computational Linguistics.

\bibitem[{Zhang et~al.(2020)Zhang, Sun, Galley, Chen, Brockett, Gao, Gao, Liu,
  and Dolan}]{zhang-etal-2020-dialogpt}
Yizhe Zhang, Siqi Sun, Michel Galley, Yen-Chun Chen, Chris Brockett, Xiang Gao,
  Jianfeng Gao, Jingjing Liu, and Bill Dolan. 2020.
\newblock \href {https://doi.org/10.18653/v1/2020.acl-demos.30} {{DIALOGPT} :
  Large-scale generative pre-training for conversational response generation}.
\newblock In \emph{Proceedings of the 58th Annual Meeting of the Association
  for Computational Linguistics: System Demonstrations}, pages 270--278,
  Online. Association for Computational Linguistics.

\end{thebibliography}
\bibliographystyle{acl_natbib}

\end{document}